\documentclass[10pt]{article}

\usepackage[a4paper,textwidth=18cm,textheight=24cm,top=2.85cm, bottom=2.85cm, left=1.5cm, right=1.5cm]{geometry}

\usepackage{icdp2009}

\makeatletter
\long\def\@makecaption#1#2{%
	\vskip\abovecaptionskip
	\sbox\@tempboxa{#1. #2}%
	\ifdim \wd\@tempboxa >\hsize
	#1. #2\par
	\else
	\global \@minipagefalse
	\hb@xt@\hsize{\box\@tempboxa\hfil}%
	\fi
	\vskip\belowcaptionskip}
\makeatother

\newcommand{\degree}{$^o$}

\usepackage{lmodern}
\usepackage{lipsum}
\usepackage{graphicx}
\usepackage{times}
\usepackage[nolist,nohyperlinks]{acronym}
\usepackage{eurosym}

\begin{document}
\noindent

\bibliographystyle{plain}

\title{A Solution for Crime Scene Reconstruction\\using Time-of-Flight Cameras}

\authorname{
Silvio Giancola$^1$,
Daniele Piron$^2$, 
Pasquale Poppa$^3$, 
Remo Sala$^4$}

\authoraddr{$^{1,2,4}$VBLAB, Vision Bricks Laboratory, Dipartimento di Meccanica, Politecnico di Milano, Italy}
\secondauthoraddr{$^3$LABANOF, Laboratorio di Antropologia e Odontologia Forense, Universita' degli Studi di Milano, Italy }

\maketitle
\footnotetext[1]{silvio.giancola@polimi.it \textit{(corresponding author)}}
\footnotetext[2]{daniele.piron@mail.polimi.it}
\footnotetext[3]{pasquale.poppa@unimi.it}
\footnotetext[4]{remo.sala@polimi.it}
\keywords
Crime Scene Reconstruction, RGB-D Camera, Kinect V2, SLAM

\abstract
In this work, we propose a method for \ac{3D} reconstruction of wide crime scene, based on a \ac{SLAM} approach. 
We used a Kinect V2 \ac{TOF} RGB-D camera to provide colored dense point clouds at a 30~Hz frequency. 
This device is moved freely (6 degrees of freedom) during the scene exploration. 
The implemented \ac{SLAM} solution aligns successive point clouds using an 3D keypoints description and matching approach.
This type of approach exploits both colorimetric and geometrical information, and permits reconstruction under poor illumination conditions. 
Our solution has been tested for indoor crime scene and outdoor archaeological site reconstruction, returning a mean error around one centimeter.
It is less precise than environmental laser scanner solution, but more practical and portable as well as less cumbersome. 
Also, the hardware is definitively cheaper.

\acresetall

\section{Introduction}
Freezing a crime scene is definitely one of the first, most important and delicate operations that the federal police must put in place as soon they arrive on a crime scene. 
The idea is to crystallize the position of objects, in order to preserve the crime scene from an inevitable contamination and, above all, to be able to keep track of all the data they will need while rebuilding the crime scene as faithfully as possible.
From a geometrical point of view, topographic surveys can solve this problematic.

In most of the current cases, the investigation is realized in the simplest way, i.e. with paper, pencil and tape measurement, drawing sketches to illustrate the environment, show the position of the victims, objects or any other material.
From the dictum of the referee that surveys the scene, someone can freely interpret what really happened, but also misunderstand it or miss a part.

As an upgrade to a simple bi-dimensional drawing someone can draw on paper, a full \ac{3D} virtual representation permits a better understanding of the evolution of certain events.
Scene reconstruction is not left to someone imagination, but can rely on actual measurements of the surrounding environment, and eventually on its dynamic.
With a consistent \ac{3D} representation, physics laws can be applied, bullets trajectories reconstruction made possible, as well as blood pattern analysis.
Is it possible to move around the virtual scene looking for new points of view without interfering with the real scene, looking for a hypothetical escape routes for a murderer.
To do so, the virtual reconstruction should not exceed a couple of centimeters uncertainty.


In this paper, we propose a solution for \ac{3D} dense reconstruction based on a \ac{TOF} camera. 
We used the Microsoft Kinect~V2 sensor, that present state-of-the-art performances for RGB-D cameras, at a cost bellow 200~\euro..
RGB-D cameras measure portions of scene or object, providing reliable colored point clouds.

\ac{SLAM} techniques consists in progressively map an unknown scene map acquiring successive frames moving freely around the scene.
The point clouds are elaborated, robust keypoints are automatically detected on the multiple measurements of the scene, matched together and aligned in an on-line manner.
Usually, keypoints are found in a \ac{2D} manner, using depth or color frame frames. 
In this work, we investigate the use of \ac{3D} keypoints described with colorimetric and geometric features.

The paper is organized in the following way. Section \ref{ch:State-of-the-Art} presents the current state of the art for \ac{3D} reconstruction applied to crime scene. In section \ref{ch:RGB-D SLAM} we present our RGB-D \ac{SLAM} pipeline based on \ac{3D} keypoints. The experimental setup is presented in section \ref{ch:setup} while results are analyzed in section \ref{ch:analysis}.

\section{State-of-the-Art}
\label{ch:State-of-the-Art}

First tentatives to reconstruct \ac{3D} crime scene with computer vision techniques can be found in~\cite{gibson2000interactive,gibson2003interactive}, where Gibson et al. used a moving camera in a \ac{SFM} pipeline.

Sansoni et al.~\cite{sansoni2009state} presented a general-purposes comparison of 3D imaging sensors for criminal investigation, taking into account most of the 3D techniques available at this time. 
It is worth noting that \ac{TOF} cameras weren't considered since the technology broadens in the very last years. 

From over a decade, \ac{SLAM} has been a hot topic in robotics, used for solving the ego-motion problem in unknown environment~\cite{durrant2006simultaneous}. 
Following this trend, Se et al.~\cite{se2005instant,se2006photo,se2008stereo} introduced the use of a two camera in a stereo SLAM pipeline, applied for crime scene reconstruction. 
Stereo mapping is a passive technique that reconstructs a 3D point cloud by matching homogeneous keypoints found on both camera.
Since it relies only on textures found on the environment, it is subject to noise where the environment lack of keypoints.
Typically, uniform indoor walls represent a bottleneck for stereo matching since no homogeneous features is present in such portion of scene. 
Similar use of stereo \ac{SLAM} for crime scene representation can be found in~\cite{kwietnewski2007multimedia}.

RGB-D \ac{SLAM} focuses on the use of depth sensors in \ac{SLAM} pipelines. 
Multiple projects~\cite{dupuis2014multi,gonzalez2015application} focused on the first version of the Kinect (Kinect~V1) for crime scene representation and object reconstruction.
Kinect~V1 is an RGB-D camera based on a structured-light technology that returns a dense colored point cloud. 
Nevertheless, structured-light sensors do not perform as well as \ac{TOF} sensors in outdoor~\cite{zennaro2015performance}. 
In fact, the natural light interfere with the infrared pattern projected by the sensor, and impede a proper depth measurement.

The updated version of the Kinect (Kinect~V2) RGB-D camera is based on the \ac{TOF} technology and then is more reliable in outdoor environment.
Moreover, using RGB-D cameras offer the possibility to use other kind of keypoints to be more reliable and independent on colors, based on geometrical features.


In this work we propose an RGB-D \ac{SLAM} pipeline using the Kinect~V2 camera that exploit \ac{3D} features for keypoints matching.

\section{Proposed approach}
\label{ch:RGB-D SLAM}
\ac{SLAM} originally focused on RGB camera. 
Its potentiality extended with the broadening of RGB-D camera.
For the RGB-D camera, we focused on the Kinect~V2 \ac{TOF} camera, while for the RGB-D SLAM we investigate the use of 3D keypoints as an extension from common RGB-D \ac{SLAM} pipelines.

\subsection{Kinect~V2}
Kinect~V2 is an RGB-D camera based on the \ac{TOF} principle. 
A pulsed \ac{IR} light is emitted from 3 LEDs, enlightens a field of view that reflects back the light measured by a matrix sensor. 
Knowing the propagation of such light in the surrounding environment, the delay between the emitted signal and the reflected one measured for every single pixel provides a depth map estimation.

\begin{figure}
    \centering
    \includegraphics[width=8.8cm]{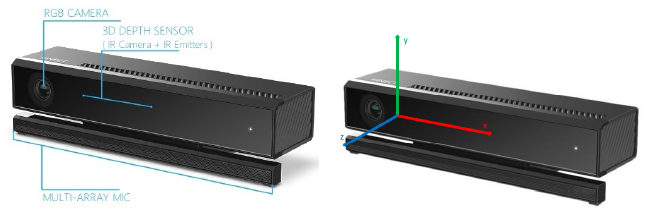}
    \caption{Kinect~V2 RGB-D camera.} 
    \label{fig:KinectV2}
\end{figure}

Kinect~V2 has a 512x424~pixel resolution sensor that snaps colored point cloud at a 30~Hz frequency up to 4.5~m. 
Note that the range can be extended up to 12~m with the libfreenect2 open-source library but uncertainty increase drastically.
Figure~\ref{fig:KinectV2} shows the sensor and Table~\ref{tab:KinectV2} summarizes its main characteristics.

Previous study focused on the characterization of such sensor~\cite{corti2015metrological}, being less than 2~mm in best case usage, growing up linearly with depth and exponentially with distance from optical axis.
Also, it is worth noting that the sensor is sensible to temperature, targets color and material and multiple path reflections.
The latter represents the main source of uncertainty, and may cause problem in finding robust 3D keypoints.
In order to be portable and independent to electrical power, a 12V car battery can substitute the 100-240~V sector transformer.

\begin{table}
    \centering
	\begin{tabular}{|lc|c|}
        \hline
		IR Camera Resolution    & [pix]	& $512$x$424$ \\
		RGB Camera Resolution   & [pix]	& $1080$x$1920$ \\
		Maximum frame rate 		& [Hz]	& $30$ \\
        Field of View 	        & [\degree]& $70(H)$x$60(V)$ \\
		Distance range  		& [mm]	& $500$x$4500$ \\
		Dimension	        	& [mm]	& $250$x$66$x$67$ \\
		Weight           		& [g]	& $966$\\
		Connection      		& 	    & USB 3.0 \\
        \hline
	\end{tabular}
    \caption{Kinect V2 for Windows - Main characteristics.}
	\label{tab:KinectV2}
\end{table}

\subsection{3D Keypoints for Simultaneous Localization and Mapping}

\ac{SLAM} focuses on aligning successive point clouds in order to reconstruct large scale environment. 
Aligning point clouds consists in finding the $SE(3)$ transformation of the current position of the camera respect to a reference one. 
The SLAM pipeline can be split in 2 phases, the front-end and the back-end.

The front-end aims to register a newly acquired point cloud with the previous one, in an on-line elaboration pipeline.
The first step consists in finding a robust sub-sample of scene points from both point clouds, describing them according to some local features, and finding the corresponding ones between 2 acquisition. 
Then, the distances of homogeneous points are minimized in 2D space~\cite{gao2003complete,lepetit2009epnp} or 3D space~\cite{besl1992method,chen1999ransac} and return the $SE(3)$ transformation that represent the pose of the camera.
The back-end focuses on finding loop closure between poses set on a graph, in order to increase the number of constraints and minimize a global residual~\cite{grisetti2010tutorial,dellaert2012factor}.
This second step is a refinement that improves large scale reconstruction.
In our approach, we will focus on the front-end part.



For our approach, we propose to use of \ac{BRISK} detection algorithm~\cite{leutenegger2011brisk} on the depth frame of the Kinect~V2.
In that way, we rely on 3D information of the acquired portion of scene. 
BRISK detection is dramatically faster than SIFT and SURF and provides similar performance~\cite{leutenegger2011brisk}.
Successively, we describe those keypoint with a \ac{C-SHOT}~\cite{tombari2011combined}.
\ac{C-SHOT} exploits both the signature efficient description with the histogram comparison robustness. 
Moreover, \ac{C-SHOT} takes into consideration both the colorimetric and depth information for describing the keypoints.
Homogeneous point are found considering most similar points in the \ac{C-SHOT} space.
Finally, we are using an \ac{ICP} algorithm~\cite{besl1992method} to find the transformation between successive point cloud considering the point-to-plane metrics~\cite{low2004linear}.

\begin{figure}[t]
    \centering
    \includegraphics[width=8.8cm]{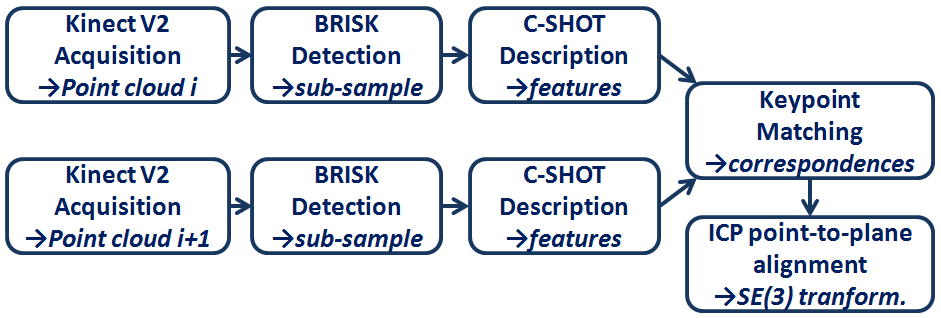}
    \caption{Our proposed registration pipeline.} 
    \label{fig:RegistrationPipeline}
\end{figure}










\section{Experimental Setup}
\label{ch:setup}
\begin{figure}[ht]
    \centering
    \includegraphics[width=8.8cm]{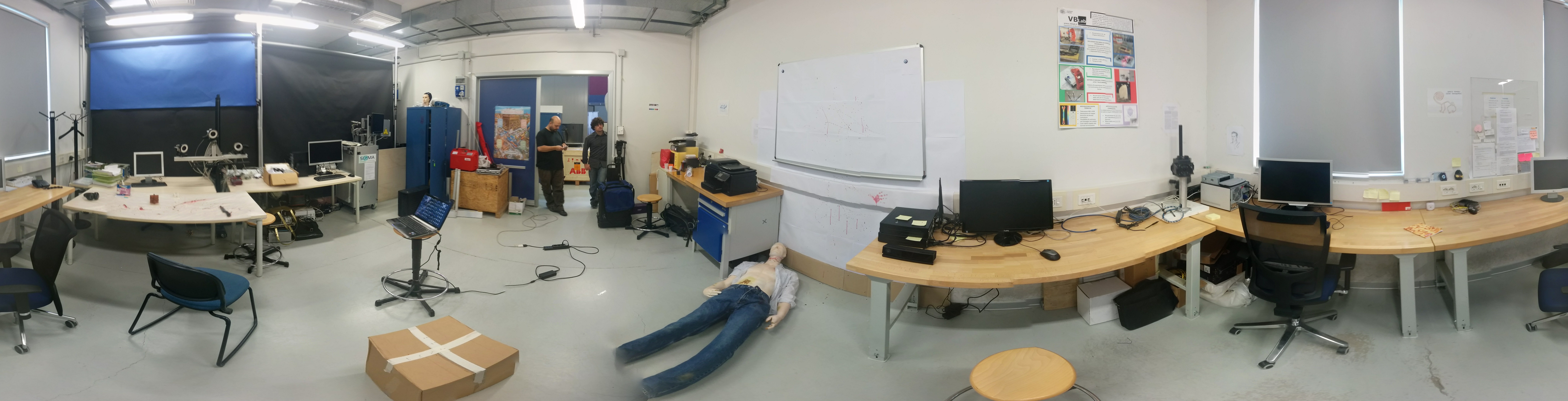}
    \caption{360\degree~photo af the simulated crime scene.} 
    \label{fig:CrimeScene360}
\end{figure}

In order to estimate the performances of the reconstruction of our SLAM pipeline, we simulate a crime scene in the \ac{VBLAB}.
We placed a mannequin in the room with evidences sparse in the scene.
Figure~\ref{fig:CrimeScene360} show a 360\degree~photo of the laboratory.
The main goal of this test is to verify the reliability of the algorithm to reconstruct the full scene, how is it possible to focus on hidden items around the scene, the performance in the darkness and finally how it respond in outdoor environment. 

\subsection{Full Scene Reconstruction}
\begin{figure}[p]
    \centering
    \includegraphics[width=8.8cm]{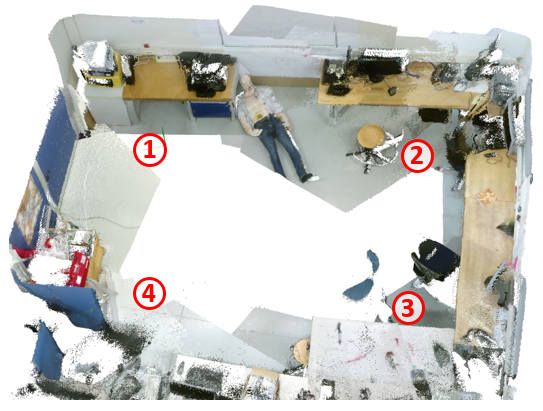}
    \caption{Reconstruction of surrounding environment.} 
    \label{fig:FullReconstruction}
\end{figure}

Regarding the full scene reconstruction it is important that this approach allow a complete scene and do not omit details.
Figure~\ref{fig:FullReconstruction} shows the complete reconstruction made from multiple acquisition that results as a patchwork of aligned point cloud. 
The corner notation from 1 to 4 will be used in the successive analysis.
We can recognize the laboratory environment and distinguish the body on the floor.
Note that the \ac{C-SHOT} description and matching is the bottleneck of our front-end \ac{SLAM} approaches since it implies a few seconds per point cloud with our configuration.

\begin{figure}[p]
    \centering
    \includegraphics[width=8.8cm]{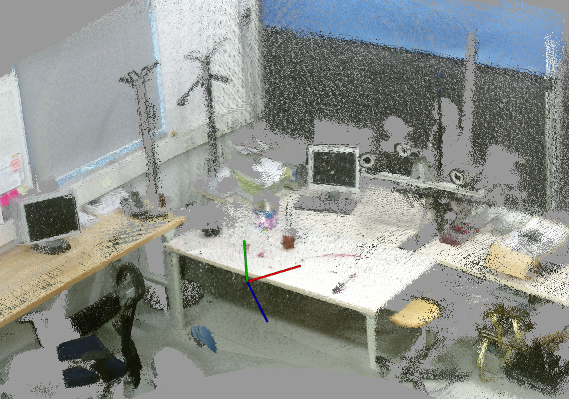}
    \caption{Focus on corner 3 part of the reconstruction.} 
    \label{fig:FocusReconstruction}
\end{figure}

\begin{figure}[p]
    \centering
    \includegraphics[width=8.8cm]{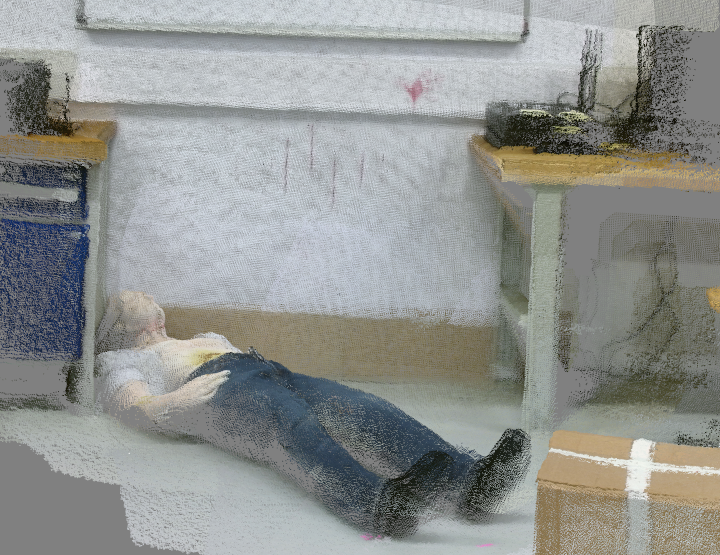}
    \caption{Focus on the mannequin in the scene.}
    \label{fig:Mannequin}
\end{figure}

\subsection{Focus on details}
\begin{figure}[p]
    \centering
    \includegraphics[height=3.3cm]{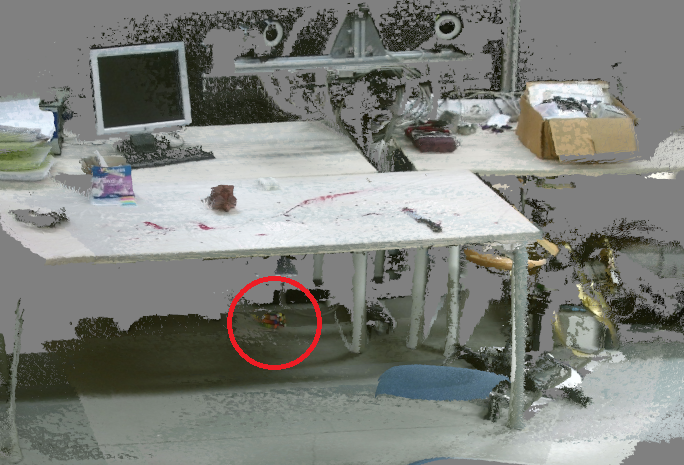}
    \includegraphics[height=3.3cm]{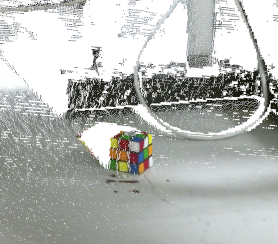}
    \caption{Rubik's cube found on corner number 3.}
    \label{fig:peculiarities3}
\end{figure}

The main advantage of using \ac{SLAM} techniques is that moving around the scene, it is also possible to focus on details.
Actually, Figure~\ref{fig:FocusReconstruction} shows the scene from a different point of view, in particular the corner 3.
In Figure~\ref{fig:Mannequin}, the reconstruction concentrated on the mannequin. 
Moving around it, occluded details can be reconstruct as well.
In Figure~\ref{fig:peculiarities3} we can find a Rubik's cube under the table of corner 3.
This Rubik's cube was hidden, and hardly remarkable if acquired with standard techniques such as environmental laser scanner, due to the occlusion of the table.
The use of the a portable the Kinect~V2 camera that can move around the scene permit the focus on details without additional effort.

\subsection{Reconstruction in the darkness}
\begin{figure}[p]
    \centering
    \includegraphics[width=8.8cm]{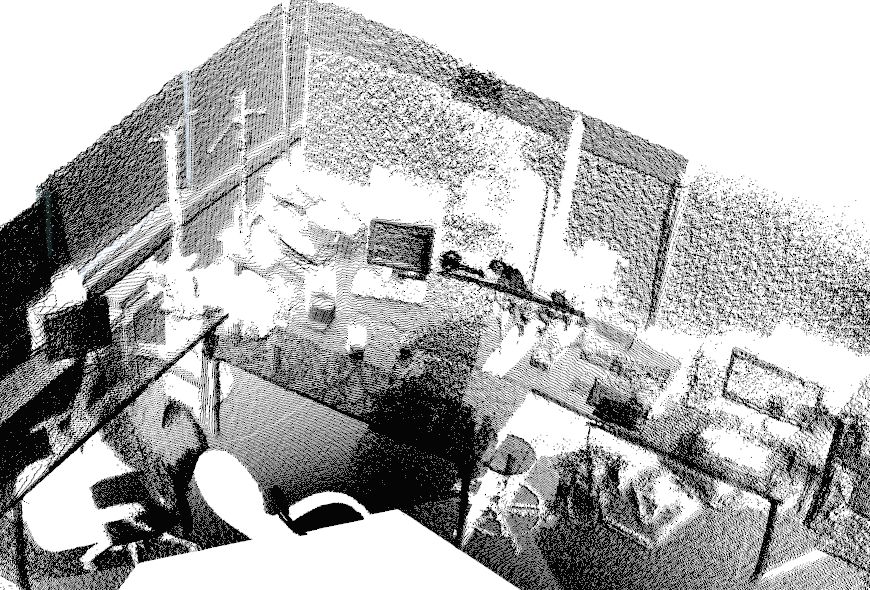}
    \caption{Reconstruction in the darkness, focused on corner 3}
    \label{fig:Darkness}
\end{figure}

One of the advantages of operating with geometrical features and do not rely only on colorimetric features is that reconstruction is possible in dark places.
In order to verify this hypothesis, we tried to reconstruct the same previous scene in the darkness.
Figure~\ref{fig:Darkness} presents the reconstruction of the corner 3.
As we can see, the reconstruction is coherent with the reconstruction shown in Figure~\ref{fig:FocusReconstruction}, without the color information. 
The depth camera do not fail in measuring the depth frame since the \ac{IR} emission do not interfere in dark places.
Also, our \ac{SLAM} approach do not lack of information for aligning successive point clouds and reconstructing the environment.
\ac{BRISK} sub-sampling is done on the depth frame while description for matching succeed using the geometrical part of the \ac{C-SHOT} description.



\subsection{Outdoor reconstruction}
\begin{figure}[p]
    \centering
    \includegraphics[width=8.8cm]{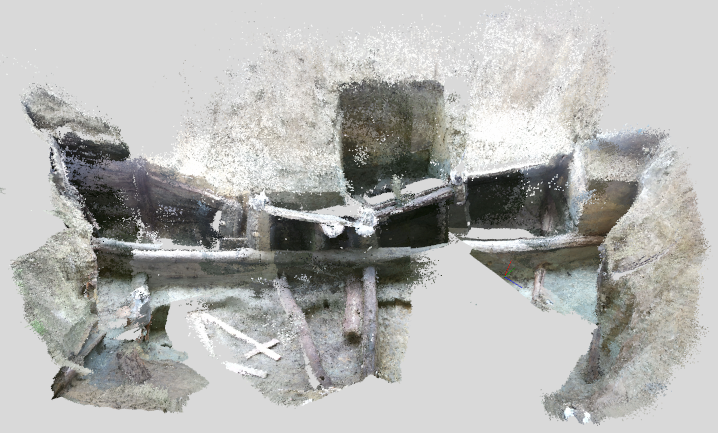}
    \caption{Global view of the archaeological outdoor reconstruction with out pipeline} 
     \label{fig:OutdoorGlobal}
\end{figure}

\begin{figure}[p]
    \centering
    \includegraphics[height=4.1cm]{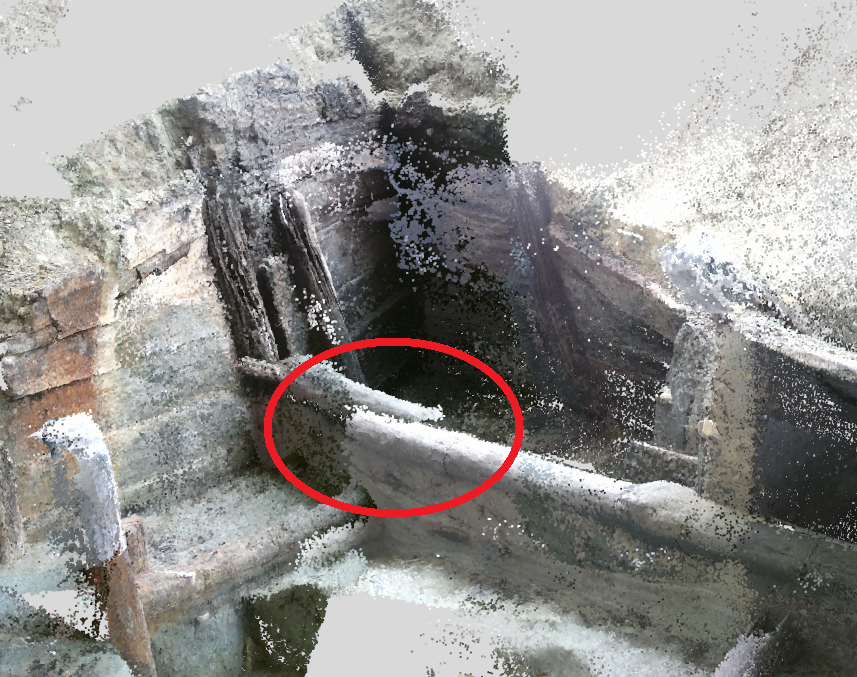}
    \includegraphics[height=4.1cm]{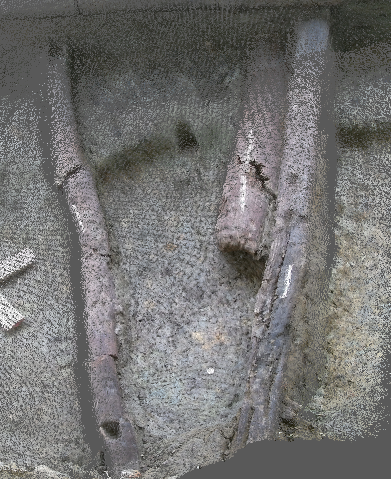}   
    \caption{Focus on the left and central part of the archaeological outdoor reconstruction} 
     \label{fig:OutdoorFocus}
\end{figure}

Regarding the outdoor reconstruction performances of the Kinect~V2, we expect that the \ac{TOF} technology handles natural light condition.
In order to stress our system, we have reconstructed an archaeological site in Noceto (PR), Italy.
It is a 500-years old pool, with $9x3x2$~m dimension.
The site is covered by a large transparent tent, that let a subdued sunny light on it. 
Figure~\ref{fig:OutdoorGlobal} presents a global view of the reconstruction made by our software.
Even if the representation looks good, the right part of the reconstruction is missing due to a track lost during the reconstruction.
Losing track occurs when the keypoints density drops.
Figure~\ref{fig:OutdoorFocus}(left) focus on the reconstruction of the central part of the site, representing the foundation.
Also, Figure~\ref{fig:OutdoorFocus}(right) enlightens an registration errors that occurred due to a loop closure not minimized.
Nevertheless, the hand-held Kinect~V2 permits the overall reconstruction moving around areas of interest of the site, focusing on desired details, and acquiring hidden and occluded parts.

\section{Analysis}
\label{ch:analysis}
In order to qualify the reconstruction the consistency of the reconstruction has been geometrically evaluated.
A consideration is done in term of error propagation and loop closure after the back-end \ac{SLAM} optimization.

\newpage

\subsection{Geometrical consistency}
Regarding geometrical consistency, the planarity and the perpendicularity of the reconstructed wall is evaluated and compared to a static reconstruction done with an environmental laser scanner.
The Stonex X300 sensor is used as ground truth reconstruction and returns a depth measurement uncertainty of a $6~mm$.
Note that moving around the environmental laser scanner is an arduous task due to its encumbrance and multiple acquisition from different point of view may be necessary if details is required.
Nevertheless, it provides a good reference for our reconstruction.

\begin{figure}[t]
    \centering
    \includegraphics[width=8.8cm]{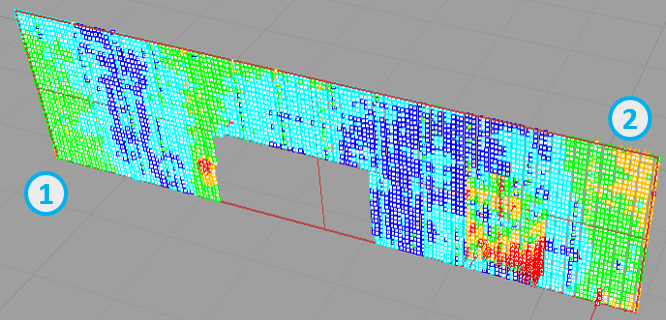}\\
    \includegraphics[width=8.8cm]{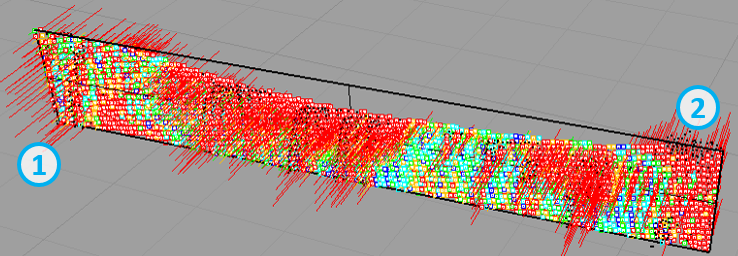}\\
    \includegraphics[width=6cm]{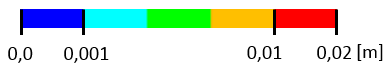}
    \caption{Planarity study of the reconstructed wall between corner 1 and 2 with Stonex X300~(top) and our SLAM solution~(bottom).} 
    \label{fig:Planarity}
\end{figure}

In Figure~\ref{fig:Planarity}, a plane is fitted with the acquisition on the wall between corner 1 and 2. 
For each point, the distance to this plane is plotted.
As expected, the Stonex provide a better reconstruction than our solution.
The Stonex hardware returns a mean error of 3.04~mm, while our solution 10.80~mm.
In both cases, the accuracy is below the tens of centimeters uncertainty required.


\begin{figure}[t]
    \centering
    \includegraphics[height=3.6cm]{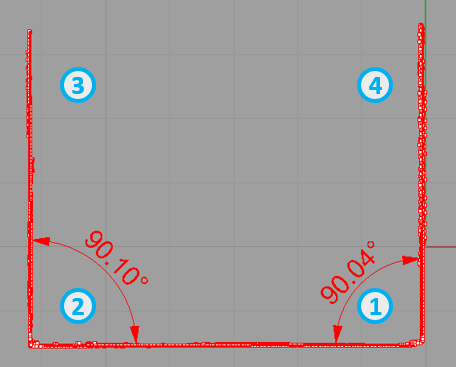}
    \includegraphics[height=3.6cm]{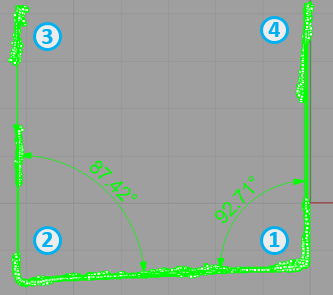}
    \caption{Perpendicularity consistency study for Stonex X300~(left) and our SLAM solution~(right).} 
    \label{fig:Perpendicularity} 
\end{figure}

In order to identify an eventual drift in the point cloud alignment which engenders an imprecise reconstruction, the perpendicularity of the reconstructed walls has also been investigated on corners 1 and 2 and compared to the Stonex reconstruction~\ref{fig:Perpendicularity}.
An offset of up to 3 degrees has been measured between the best fitting planes representing the walls for our pipeline reconstruction, while this imprecision is negligible in the Stonex reconstruction.

For a general purposes scene interpretation, it can suffice to understand broadly the environment, with an uncertainty over the centimeter in position and of a few degree for the registration orientation.
Moreover, if portability and practicality is required, the Kinect~V2 is the most suitable solution since it can reach spots that are hardly reachable with more precise laser scanner systems.


\subsection{Propagation error and loop closure optimization}

As we hypothesize, SLAM techniques introduce a propagation error non negligible, but robust loop closure detection can reduce this drift.
We actually noticed that it is not always the case.
When a bad registration due to a bad correspondences matching is performed, the overall reconstruction is not reliable anymore.
The reconstruction error increases with the distance and the misalignment is tricky to correct.

\begin{figure}[t]
    \centering
    \includegraphics[width=2.8cm]{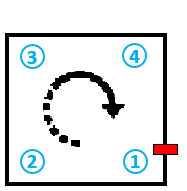}
    \includegraphics[width=2.8cm]{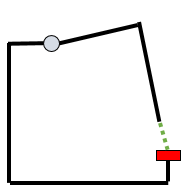}
    \includegraphics[width=2.8cm]{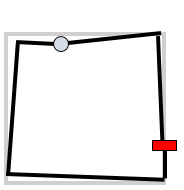}
    \caption{Loop closure~(red rectangle) after a bad registration~(grey circle).} 
    \label{fig:LoopClosure} 
\end{figure}

This problem is illustrated in Figure~\ref{fig:LoopClosure}, a reconstruction occurs starting from corner 1 to corner 4 in a clockwise rotation (left part of the figure).
Between corner 3 and 4, one registration is badly performed (grey circle in the middle), due to a bad correspondences matching.
Adding a loop closure between first and last frame, the overall residual should minimize, but not the erroneous registration.
In that case, the global minimization tends to distort the complete reconstruction.




\section{Conclusion}
\label{ch:conclusion}


In this project we presented a RGB-D \ac{SLAM} pipeline for crime scene reconstruction.
We used a Kinect~V2 \ac{TOF} camera in order to acquire dense portions of the scene.
We successively detect \ac{BRISK} keypoints on depth frames that are described with \ac{C-SHOT} colorimetric and geometric features.
Not relying only on colorimetric information permits reconstruction in poor light condition.
Also, using a \ac{TOF} camera for depth sensing improve dramatically measurement reliability in natural outdoor lightening condition.

The proposed approach has been tested in two real case applications, a simulated crime scene and a archaeological site.
Results show that our approach is more practical and less cumbersome than environmental laser scanner solution, as well as less expensive.
Also it fulfills the application requirement, providing a single-centimeter reconstruction accuracy.

The current approach is limited by the elaboration time of the \ac{C-SHOT} description and matching. Other descriptor could be compared such as \textit{SIFT3D} and the \textit{BRISK} descriptor, in terms of performance reliability and timing.
Future works include a comparison of reconstruction performances of other sensors such as the \textit{Sense} depth camera based on structured-light and the \textit{Minolta Vivid 910} laser scanner.

\begin{acronym}[OpenFOAM]	
    \acro{2D}{bi-dimensional}
    \acro{3D}{three-dimensional}
    \acro{BRISK}{Binary Robust Invariant Scalable Keypoints}
    \acro{C-SHOT}{Colored Signature of Histograms of OrienTations}
    \acro{ICP}{Iterative Closest Point}
    \acro{IR}{Infra-red}
    \acro{RANSAC}{Random Sample Consensus}
    \acro{RTABMAP}{Real-Time Appearance Based Mapping}
    \acro{SFM}{Structure-from-Motion}
    \acro{SHOT}{Signature of Histograms of OrienTations}
    \acro{SLAM}{Simultaneous Localization and Mapping}
    \acro{TOF}{Time-of-Flight}
    \acro{VBLAB}{Vision Bricks Laboratory}
\end{acronym}

\bibliography{Bibliograghy}

\end{document}